\pdfoutput=1

\documentclass[letterpaper, 10 pt, conference]{ieeeconf}  %

\IEEEoverridecommandlockouts                              %

\usepackage{graphicx} %
\usepackage{amsmath} %
\usepackage{amssymb} %
\usepackage{amsfonts} %
\usepackage{siunitx}
\usepackage{url}
\usepackage{xspace}

\graphicspath{{images/}}
\DeclareGraphicsExtensions{.pdf,.jpeg,.png}

\newcommand{\vect}[1]{\mathbf{#1}} %
\newcommand{\field}[1]{\mathbb{#1}}
\newcommand{\conj}[1]{#1^{*}} %
\newcommand{\pre}[2]{{}_{#2}^{#1}}
\newcommand{\rot}[3]{\pre{#2}{#3}#1}
\newcommand{\axis}[3]{\pre{#1}{}\vect{#2}_{#3}}
\newcommand{\st}{:} %
\newcommand{\Hsym}{\field{H}} %
\newcommand{\I}{\field{I}} %
\newcommand{\Q}{\field{Q}} %
\newcommand{\R}{\field{R}} %
\providecommand{\abs}[1]{\lvert#1\rvert}
\providecommand{\norm}[1]{\lVert#1\rVert}
\newcommand{\fr}[1]{\{#1\}\xspace}
\newcommand{\smidge}{\mspace{1mu}}
\newcommand{\nsmidge}{\mspace{-3mu}}
\newcommand{\nop}{NimbRo\protect\nobreakdash-OP\xspace}

\newcommand{\cpp}{C\texttt{\nolinebreak\hspace{-.05em}+\nolinebreak\hspace{-.05em}+}\xspace}
\DeclareMathOperator{\acos}{acos}
\DeclareMathOperator{\tr}{tr}

\newcommand{\eqnref}[1]{(\ref{eqn:#1})\xspace} %
\newcommand{\eqnrefs}[2]{(\ref{eqn:#1}--\ref{eqn:#2})\xspace}
\newcommand{\secref}[1]{Section~\ref{sec:#1}\xspace}

\newcommand{\figref}[1]{Fig.~\ref{fig:#1}\xspace}

\title{\LARGE \bf Robust Sensor Fusion for Robot Attitude Estimation}

\author{Philipp Allgeuer and Sven Behnke%
\thanks{All authors are with the Autonomous Intelligent Systems (AIS) Group, Computer Science Institute VI,
        University of Bonn, Germany. Email: {\tt\small pallgeuer@ais.uni-bonn.de}. This work was partially
        funded by grant BE 2556/10 of the German Research Foundation (DFG).}}

\usepackage{eso-pic}

\AtBeginDocument{\AddToShipoutPictureFG*{\AtTextUpperLeft{\put(0,\LenToUnit{9pt}){\parbox{\textwidth}{\centering\bfseries
International Conference on Humanoid Robots (Humanoids), Madrid, Spain, 2014
}}}}}
\begin{document}

\maketitle
\thispagestyle{empty}
\pagestyle{empty}

\begin{abstract}
Knowledge of how a body is oriented relative to the world is frequently
invaluable information in the field of robotics. An attitude estimator that
fuses 3-axis gyroscope, accelerometer and magnetometer data into a quaternion
orientation estimate is presented in this paper. The concept of fused yaw, used
by the estimator, is also introduced. The estimator, a nonlinear complementary
filter at heart, is designed to be uniformly robust and stable---independent of
the absolute orientation of the body---and has been implemented and released as
a cross-platform open source \cpp library. Extensions to the estimator, such as
quick learning and the ability to deal dynamically with cases of reduced sensory
information, are also presented.
\end{abstract}

\section{Introduction}

Attitude estimation is the task of constructing an estimate of the full 3D
orientation of a body relative to some global fixed frame, based on a finite
history of sensor measurements. The body in question is often a robot, but in
total generality it can correspond to any object that is equipped with the
sensors necessary for the estimation task. With the advent of low cost inertial
sensors---particularly those based on microelectromechanical systems
(MEMS)---the field of application for attitude estimation techniques has greatly
widened, extending into the field of low cost robotics. With low cost sensors
and processors however, it is crucial that any estimation algorithms are able to
run computationally efficiently, and are able to function with high noise inputs
without excessively sacrificing estimator response. In addition to low estimator
latency, orientation-independent mathematical and numerical stability is also
desirable.

An attitude estimator that aims to fulfil the aforementioned criteria is
presented in this paper. The estimator has been implemented as a generic
portable \cpp library, and is freely available online \cite{AttEstGithub}. All
of the algorithms and cases discussed in this paper are implemented in the
release, and have been tested both in simulation and on a real humanoid
platform, the \nop \cite{Allgeuer2013a}, developed by the University of Bonn.

Much effort has been made in the past to develop algorithms for the
reconstruction of attitude in aeronautical environments. This work was largely
in relation to the attitude and heading reference systems (AHRS) required for
aeronautical applications, with examples being the works of Gebre-Egziabher et
al. \cite{Gebre2004} and Mungu\'ia and Grau \cite{Munguia2014}. Other works in
the area of attitude estimation, such as Vaganay et al. \cite{Vaganay1993} and
Balaram \cite{Balaram2000}, have focused more on robotics and control
applications, but do not specifically address the issues encountered with low
cost inertial measurement unit (IMU) systems. A comprehensive survey of modern
nonlinear filtering methods for attitude estimation was undertaken by Crassidis
et al. \cite{Crassidis2007}. Almost all of the surveyed advanced filtering
techniques relied on some form of the Extended Kalman Filter (EKF), with various
modifications being used to improve particular characteristics of the
filter---often convergence. Such EKF filters can be seen to be computationally
expensive however, when considering implementation on embedded targets such as
microcontrollers. It is often also difficult to provide a guarantee of filter
robustness \cite{Euston2008}.

Alternative to the general stream of development of EKF filtering is the concept
of complementary filtering. This builds on the well-known linear single-input
single-output (SISO) complementary filters, and extends these in a nonlinear
fashion to the full 3D orientation space. Such filters have favourable frequency
response characteristics, and seek to fuse low frequency attitude information
with high frequency attitude rate data. Prominent examples of generalised
complementary filtering include the works of Jensen \cite{Jensen2011} and Mahony
et al. \cite{Mahony2008}.

The problem addressed in this paper relates specifically to the design of an
attitude estimator that can function with noisy low cost sensors and is simple
and efficient enough to be implemented at high loop rates on low power embedded
targets, such as microcontrollers. To this end, the work presented by Mahony et
al. in \cite{Mahony2008} was used as a basis for the attitude estimator
developed in this paper. A central problem in applying this work however, is
that a method is required for reconstructing an instantaneous 3D orientation
`measurement' directly from the sensor measurements. This is a complex
optimisation problem that generally requires a suboptimal solution algorithm for
computational feasibility reasons \cite{Mahony2008}. Literature does not elucidate a
clear solution to this problem---in particular not in an explicit form---and not
in a way that can function robustly in all cases. The contribution of this paper
lies predominantly in the presentation of an algorithm for robust calculation of
such instantaneous orientation measurements. Other contributions include the
novel use of fused yaw (\secref{yawrotation}), the integration of quick learning
(\secref{quicklearning}), and the explicit extension of the attitude estimator
to cases of reduced sensory information (\secref{extensions}). A summary of the
notation and identities used in this paper is provided in the appendix.

\section{Preliminaries}

\subsection{Problem Definition}

The goal of attitude estimation is to calculate an estimate of the rotation of a
body relative to a global fixed frame, based on observations acquired through
sensory perception. Such sensory perception can include accelerometers,
gyroscopes, magnetometers, Global Positioning System (GPS), visual perception
and/or LIDAR. The types of sensors considered for the task in this paper are the
ones that are typically found in IMU systems, and typically available in low
cost variants for mobile robotic systems. These are the first three in the
preceding list. Irrespective of which sensors are used however, it is a
stringent requirement that the estimator always remains stable, and is able to
function equally well throughout the entire orientation space.

We define \fr{B} to be the body-fixed frame, which rotates with the body, and
with the sensors that provide the observational input to the attitude estimator.
It is assumed that \fr{B} is defined such that its z-axis points `upwards'
relative to the body, and its x-axis points `forwards'. We define \fr{G} to be
the global fixed frame, with the convention that the z-axis points `upwards'
relative to the world. Importantly, this means that the gravity vector can be
written as $\vect{g} = (0,0,-g)$ in global coordinates.

Using these definitions, the problem considered in this paper can be more
precisely reformulated as being the task of robustly calculating an estimate for
$\rot{q}{G}{B}$ (or $\rot{R}{G}{B}$), given arbitrary 3-axis gyroscope,
accelerometer and magnetometer data. The format and type of data provided by
each of these sensors is assumed to be modelled as follows.

\subsubsection{Gyroscope}
This sensor is assumed to provide a measure $\axis{B}{\Omega}{y} \in \R^3$ of
the angular velocity of the body, in the coordinates of frame \fr{B}. The
measurement is assumed to be affected by a largely time-invariant gyroscope bias
$\vect{b}_{\Omega}$, as well as zero mean sensor noise $\vect{v}_{\Omega}$. That is,
\begin{equation}
\axis{B}{\Omega}{y} = \axis{B}{\Omega}{} + \vect{b}_{\Omega} + \vect{v}_{\Omega} \in \R^3,
\end{equation}
where $\axis{B}{\Omega}{}$ is the true angular velocity of the body. The
measurement $\axis{B}{\Omega}{y}$ must be expressed in \si{\radian\per\second}.

\subsubsection{Accelerometer}
This sensor is assumed to provide a measure $\axis{B}{\tilde{a}}{} \in \R^3$ of
the proper acceleration of the body. This is the inertial acceleration being
experienced by the body, together with the effect of gravitational acceleration.
The latter term is assumed to dominate the measured proper acceleration. Cases
where this assumption is violated, such as in collisions, are implicitly
filtered out by the low-pass dynamics of the estimator. Merging the inertial
acceleration components into the noise term $\vect{v}_a$ gives
\begin{equation}
\axis{B}{\tilde{a}}{} = \rot{R}{B}{G}\smidge\smidge\axis{G}{g}{} + \vect{b}_a + \vect{v}_a \in \R^3, \label{eqn:acctilde}
\end{equation}
where $\axis{G}{g}{} = (0,0,-g)$ is the global gravity vector, and $\vect{b}_a$
is a time-invariant accelerometer bias. It is assumed that an estimate
$\vect{\hat{b}}_a$ of the bias is available from an accelerometer calibration,
and can be used to unbias $\axis{B}{\tilde{a}}{}$. Normalising the unbiased
$\axis{B}{\tilde{a}}{}$ with zero noise, and recalling \eqnref{rotmatrowcol}, yields
\begin{equation}
\axis{B}{z}{G} = -\frac{\axis{B}{\tilde{a}}{}-\vect{\hat{b}}_a}{\norm{\axis{B}{\tilde{a}}{}-\vect{\hat{b}}_a}} \in S^2.
\end{equation}
Thus, with the assumption that the accelerometer measurement points in the
direction of gravity, an instantaneous measurement of $\axis{B}{z}{G}$ can be
derived for use in the filter.

\subsubsection{Magnetometer}
This sensor is assumed to provide a measure $\axis{B}{\tilde{m}}{} \in \R^3$ of
the strength and direction of $\axis{G}{m}{e}$, the Earth's magnetic field, in
the coordinates of frame \fr{B}. The measurement is assumed to be affected by a
largely time invariant bias $\vect{b}_m$, induced by local magnetic
disturbances, as well as zero mean sensor noise $\vect{v}_m$. Therefore
\begin{equation}
\axis{B}{\tilde{m}}{} = \rot{R}{B}{G}\smidge\smidge\axis{G}{m}{e} + \vect{b}_m + \vect{v}_m \in \R^3,
\end{equation}
where $\rot{R}{B}{G}$ is the inverse of the current true orientation of the
body. It is the purpose of a hard-iron magnetometer calibration to derive an
estimate $\vect{\hat{b}}_m$ for $\vect{b}_m$, which is then subtracted from all
future measurements. Assuming a non-zero unbiased field strength, subsequent
normalisation yields
\begin{equation}
\axis{B}{m}{} = \frac{\axis{B}{\tilde{m}}{} - \vect{\hat{b}}_m}{\norm{\axis{B}{\tilde{m}}{} - \vect{\hat{b}}_m}}
\approx \rot{R}{B}{G} \cdot \frac{\axis{G}{m}{e}}{\norm{\axis{G}{m}{e}}} \in S^2.
\end{equation}

\subsection{Yaw of a Rotation}
\label{sec:yawrotation}

There are many different possible definitions for the \emph{yaw} of a rotation.
Most of these definitions are equivalent to the first parameter of one of the
twelve Euler angle rotation representations. Various different conventions of
Euler angles exist, some alternately referred to as Tait-Bryan angles, and as
such many different definitions of yaw can be derived. However, as yaw should
intuitively correspond to some notion of how rotated a frame is about the global
axis that points `upwards'---in this case the z-axis---the natural choice here
is the Z-Y$'$-X$''$ Euler angle convention. The Z component of the ZYX Euler
angles representation of a rotation is henceforth referred to as the
\emph{ZYX yaw} of that rotation, and denoted $\psi_E$.

Analysis of the definition of ZYX yaw leads to a particularly useful
characterisation thereof---the ZYX yaw of a rotation from frame \fr{A} to frame
\fr{B} is equivalent to the angle about $\axis{}{z}{A}$ from $\axis{}{x}{A}$ to
the projection of $\axis{}{x}{B}$ onto the $\axis{}{x}{A}\axis{}{y}{A}$ plane.
Thus, two frames \fr{B} and \fr{C} have the same ZYX yaw relative to \fr{A} if
the projections of their respective x-axes onto the $\axis{}{x}{A}\axis{}{y}{A}$
plane are parallel. Note that `parallel' here is a stronger assertion than pure
collinearity (refer to the appendix). From this characterisation it can be seen
that the ZYX yaw goes undefined when $\axis{}{x}{B}$ is collinear with
$\axis{}{z}{A}$. This corresponds to the well-known gimbal lock phenomenon, and
is a singularity of this definition of yaw. For the sample application of a
humanoid robot however, it is not uncommon that this configuration is reached,
which can be problematic depending on implementation. ZYX yaw also does not
possess some properties that can be useful in a definition of yaw, such as
negation through rotation inversion.

In light of these issues, the notion of \emph{fused yaw} is proposed as an
alternative definition of yaw. Given two frames \fr{A} and \fr{B}, in general
there is a unique rotation that maps $\axis{}{z}{B}$ onto $\axis{}{z}{A}$, such
that the axis of rotation is perpendicular to both $\axis{}{z}{B}$ and
$\axis{}{z}{A}$. \fr{C} is defined to be the frame that results from applying
this rotation to \fr{B}. The fused yaw of the rotation from \fr{A} to \fr{B},
denoted $\psi_F$, is defined as the angle from $\axis{}{x}{A}$ to
$\axis{}{x}{C}$ about $\axis{}{z}{A}$. This definition is only ambiguous if
$\axis{}{z}{A}$ is antiparallel to $\axis{}{z}{B}$. This corresponds to a point
of singularity of fused yaw---unavoidable in general definitions of yaw---and
can be thought of as the frame \fr{B} being `upside down' relative to \fr{A}.
Note that if $\axis{}{z}{A}$ and $\axis{}{z}{B}$ are parallel, then \fr{C} is
unambiguously taken to be \fr{B}, and the fused yaw is still well-defined.
Although beyond the scope of this paper, the definition of fused yaw is
consistent, well-defined, and satisfies the axiomatic conditions one would
expect of an expression of yaw. Fused yaw also has some useful properties, such
as negation through rotation inversion.

\subsection{1D Linear Complementary Filter}

A simple preliminary approach to the attitude estimation problem is to separate
the problem into each of its independent axes of rotation. This can work well
for body rotations close to the upright identity pose, but does not extend well
to the whole orientation space. Nevertheless, the 1D filtering approach
demonstrates well the concept of linear complementary filtering. Taking for
example the pitch direction of rotation, one can express the filter equations as
\begin{align}
\dot{\hat{\theta}} &= (\omega_y - \hat{b}_\omega) + k_p(\theta_y - \hat{\theta}) \label{eqn:1dangle}\\
\dot{\hat{b}}_\omega &= -k_i(\theta_y - \hat{\theta}), \label{eqn:1dbias}
\end{align}
where $\hat{\theta}$ is the pitch angle estimate, $\theta_y$ is an instantaneous
measure of the pitch angle based solely on the accelerometer, $\omega_y$ is the
gyroscope measurement in the pitch direction, $\hat{b}_\omega$ is an estimate of
the bias thereof, and $k_p$ and $k_i$ are PI compensator gains. A similar
expression can be formulated for the roll and yaw directions, where in the
latter case the $\theta_y - \hat{\theta}$ error term is left to zero. The PI
compensator closes the loop on the type~I system, forming a linear second order
system with zero theoretical steady state error to step inputs. The linear
complementary filter combines the high-pass rate data with the low-pass position
data to form a high bandwidth estimate of the system state. However, despite
possessing positive filter attributes, the assumption that each axis behaves
independently places a severe limitation on the usability of the filter for
attitude estimation. A core issue is that the angular velocity about one axis
generally affects the rotation about all axes, and to differing amounts
depending on the orientation of the body. The 1D filter also does not lead
unambiguously to some notion of a total 3D orientation.

\section{3D Nonlinear Passive Complementary Filter}

\subsection{Motivation and Filter Type}
\label{sec:motivation}

In light of the limitations of the 1D complementary filter, it is desirable to
formulate a complementary filter that operates on the full 3D rotation space,
ideally retaining the positive frequency attributes of the linear filter. Mahony
et al. \cite{Mahony2008} introduced three such nonlinear filters, the direct,
passive and explicit complementary filters. The main difference between the
three filters is that while the direct complementary filter uses the
instantaneous inertial sensor data to transform the gyroscope measurements in
the update equation, the passive complementary filter uses the current filter
estimate, and the explicit complementary filter uses an update technique that
operates directly on the sensor measurement vectors.

A key design decision of the attitude estimator presented here is that the
magnetometer measurements should not have any direct influence on the attitude
estimate, other than to resolve the yaw. The reason for this is to reduce
instabilities in the output pitch and roll components, and to alleviate the
requirement for a magnetometer calibration for these components of the estimate
to function correctly. This is not possible to achieve with the explicit
complementary filter, and so the only filter in \cite{Mahony2008} to provide a
solution to the problem of constructing an instantaneous orientation measurement
from sensor data was found to be unsuitable. Comparison of the direct and
passive filters also led to the conclusion that the feed-forward nature of the
direct formulation was unsuitable due to high frequency noise considerations.
Consequently, the attitude estimator presented in this paper was built around
the core of the nonlinear passive complementary filter.

\subsection{Passive Complementary Filter Equations}

We define frame \fr{E} as the frame corresponding to the current estimate of the
body's orientation, $\rot{\hat{q}}{G}{E} \equiv \hat{q}$. Given the current
sensor measurements $\axis{B}{z}{G}$ and $\axis{B}{m}{}$, and if needed also
$\hat{q}$, the first task is to construct a full 3D `measured' orientation
$\rot{q}{G}{B}_y \equiv q_y$ that is consistent with these measurements. Frame
\fr{B} is implicitly defined via this measured orientation. The error in the
current orientation estimate with respect to the sensor measurements is
expressed as $\tilde{q} = \hat{q}^* q_y$, where $\tilde{q} \equiv
(\tilde{q}_0,\vect{\tilde{q}}) \equiv \rot{\tilde{q}}{E}{B}$. The axis of this
rotation leads to the corrective error feedback term $\vect{\Omega}_e$, which is
added to the unbiased measured angular velocity using the equations
\begin{align}
\vect{\Omega}_e &= 2\tilde{q}_0\vect{\tilde{q}} \label{eqn:3derror}\\
\vect{\Omega} &= \vect{\Omega}_y - \hat{\vect{b}}_\Omega + k_p\vect{\Omega}_e,
\end{align}
where $\hat{\vect{b}}_\Omega$ is the current estimate of the gyroscope bias, and
$k_p$ is a P gain. The filter equations are then
\begin{align}
\dot{\hat{q}} &= \tfrac{1}{2} \hat{q} \smidge\Omega \label{eqn:qhatupdate}\\
\dot{\hat{\vect{b}}}_\Omega &= -k_i \vect{\Omega}_e, \label{eqn:3dbias}
\end{align}
where $\Omega = (0,\vect{\Omega}) \in \Hsym$, and $k_i$ is an I gain. Note that
mathematically \eqnref{qhatupdate} simply converts the angular velocity $\Omega$
into a quaternion angular velocity that can then be integrated. Trapezoidal
integration is recommended for numerical implementations of these equations. We
recommend that the time increment used for the numerical integration be the
measured time, coerced to a suitable range, such as $[0.8,2.2]$ times the
nominal update interval of the filter. This avoids large jumps in the estimator
states when lags occur and ensures greater correctness of the gyroscope
integration, leading to better estimation results. The PI gains of the filter
should be tuned to provide non-oscillatory yet responsive transients, as limited
by sensor noise. The similarities between the passive filter and the 1D
complementary filter become apparent when comparing \eqnrefs{1dangle}{1dbias}
and \eqnrefs{3derror}{3dbias}.

The stability of the passive complementary filter is discussed in detail in
\cite{Mahony2008}. Theoretical analysis demonstrates that there is a measure
zero set in the space of all possible measured rotation and bias errors such
that equilibrium exists despite lack of convergence. The equilibrium is unstable
however, and the error is locally exponentially stable in all other cases. This
set consists of all error states such that $\hat{\vect{b}}_\Omega$ is
error-free, and $\tilde{q}$ is a rotation by $\pi$ radians. This pathological
set is of no concern however, as it is never reached in any practical situation.
Even intentional initialisation of the filter to such an equilibrium state in
simulated experiments did not prove to be a problem, as mere arithmetic floating
point errors were enough for the divergent dynamics near the pathological set to
take over.

\section{Measured Quaternion Orientation Resolution}

\subsection{General Case}

The calculation of the estimation error quaternion $\tilde{q}$, requires
knowledge of $q_y$, the instantaneous measured orientation best fitting the
sensor measurements $\axis{B}{z}{G}$ and $\axis{B}{m}{}$. In general, these two
measurements suffice to construct a unique rotation $q_y$ that best fits the
given data. If not, $\hat{q}$ is taken as a further input, and one of the
\emph{resolution methods} described in the following sections is used. In
absolutely all cases however, $q_y \equiv \rot{q}{G}{B}_y$ must respect
$\axis{B}{z}{G}$. This ensures that the magnetometer, and any assumptions made
by the resolution methods, as desired do not affect the pitch and roll
components of the output quaternion estimate.

In the general case, a value for $\axis{G}{m}{e}$ is required. This is easily
obtained by physically rotating the body such that the \fr{B} and \fr{G} frames
coincide, and setting $\axis{G}{m}{e}$ to $\axis{B}{m}{}$. This vector is only
used as a reference relative to which the yaw of the output quaternion is
expressed. The goal is to find a suitable rotation matrix
$\rot{R}{G}{B}_y = \begin{bmatrix} \axis{B}{x}{G} & \!\!\!\axis{B}{y}{G}\!\!\!\! & \axis{B}{z}{G} \end{bmatrix}^T$,
and convert it into the required quaternion $\rot{q}{G}{B}_y$. Refer to the
appendix for details of a known robust conversion algorithm.

Ideally we would wish to be able to find $\axis{B}{x}{G}$ and $\axis{B}{y}{G}$
such that $\axis{G}{m}{e}$ and $\axis{B}{m}{}$ are equal, but as this is not
necessarily possible, we instead minimise the angular difference between the
two. This condition can be seen to be satisfied when the respective projections
of the two vectors onto the plane perpendicular to $\axis{B}{z}{G}$ are
parallel. We define $\axis{B}{\hat{m}}{}$ to be the projection of
$\axis{B}{m}{}$ onto the $\axis{}{x}{G}\axis{}{y}{G}$ plane, and use the cross
product to construct a suitable third basis vector $\axis{B}{\hat{u}}{}$. The
required $\axis{B}{x}{G}$ and $\axis{B}{y}{G}$ vectors are then calculated as
linear combinations of these basis vectors, based on the condition that
$\axis{B}{\hat{m}}{}$ must be parallel to
$\axis{G}{\hat{m}}{e} = (m_{ex},m_{ey},0)$, the trivial projection of
$\axis{G}{m}{e}$ onto the $\axis{}{x}{G}\axis{}{y}{G}$ plane. This algorithm can
be summarised mathematically as
\begin{align}
\axis{B}{\hat{m}}{} &= \axis{B}{m}{} - (\axis{B}{m}{} \cdot \axis{B}{z}{G})\axis{B}{z}{G} \label{eqn:magprojection}\\
\axis{B}{\hat{u}}{} &= \axis{B}{\hat{m}}{} \times \axis{B}{z}{G} \\
\axis{B}{\tilde{x}}{G} &= m_{ex}\axis{B}{\hat{m}}{} + m_{ey}\axis{B}{\hat{u}}{}\\
\axis{B}{\tilde{y}}{G} &= m_{ey}\axis{B}{\hat{m}}{} - m_{ex}\axis{B}{\hat{u}}{}\\
\axis{B}{x}{G} &= \frac{\axis{B}{\tilde{x}}{G}}{\norm{\axis{B}{\tilde{x}}{G}}},\;\;
\axis{B}{y}{G} = \frac{\axis{B}{\tilde{y}}{G}}{\norm{\axis{B}{\tilde{y}}{G}}} \label{eqn:normalisevecs}\\
\rot{R}{G}{B}_y &= \begin{bmatrix} \axis{B}{x}{G} & \axis{B}{y}{G} & \axis{B}{z}{G} \end{bmatrix}^T \label{eqn:vecstorotmat}
\end{align}
Note that $m_{ez}$ is not required. This algorithm only fails with a division by
zero if $\axis{G}{\hat{m}}{e}$ is degenerate, or $\axis{B}{z}{G}$ and
$\axis{B}{m}{}$ are collinear. Both these causes of failure correspond to
measurements of the Earth's magnetic field being vertical in the global fixed
frame, a generally unexpected case.

\subsection{ZYX Yaw Orientation Resolution Method}

If the general case fails to produce a valid output, the magnetometer
measurement is discarded, and a measured orientation $q_y$ is instead
constructed from $\axis{B}{z}{G}$ and $\hat{q}$. The latter is required as the
former alone is insufficient to be able to calculate a unique $q_y$, and the
latter can be used to ensure that $q_y$ is `as close as possible' to $\hat{q}$,
thereby only minimally affecting the estimate in uncontrolled dimensions (i.e.\ yaw).

We define the frame \fr{H} to be the frame \fr{B} rotated by the inverse of
$\hat{q}$. That is, \fr{H} corresponds to the current estimated orientation of
the global fixed frame. Note that this will not be identical to \fr{G} in
general, as $\hat{q}$ and $q_y$ generally differ, even if only slightly. The aim
of the ZYX yaw resolution method is to find a suitable rotation matrix
$\rot{R}{G}{B}_y$, such that the ZYX yaw of \fr{H} with respect to \fr{G} is
zero. This is equivalent to saying that $\axis{B}{x}{G}$ should be parallel, and
hence equal to, the normalised projection of $\axis{B}{x}{H}$ onto the
$\axis{}{x}{G}\axis{}{y}{G}$ plane. $\axis{B}{y}{G}$ is calculated to complete
the orthogonal basis. Letting $\hat{q} = (\hat{w},\hat{x},\hat{y},\hat{z})$, the
algorithm is mathematically given as
\begin{align}
\axis{B}{x}{H} &= (0.5-\hat{y}^2-\hat{z}^2,\: \hat{x}\hat{y}-\hat{w}\hat{z},\: \hat{x}\hat{z}+\hat{w}\hat{y}) \label{eqn:zyxyawmethodbegin}\\
\axis{B}{\tilde{x}}{G} &= \axis{B}{x}{H} - (\axis{B}{x}{H} \cdot \axis{B}{z}{G})\axis{B}{z}{G} \\
\axis{B}{\tilde{y}}{G} &= \axis{B}{z}{G} \times \axis{B}{\tilde{x}}{G} \label{eqn:zyxyawmethod}
\end{align}
after which equations \eqnrefs{normalisevecs}{vecstorotmat} are used as before.
The obtained rotation matrix $\rot{R}{G}{B}_y$ is converted into the required
quaternion $\rot{q}{G}{B}_y$. This algorithm only fails, in the form of a
division by zero, if $\axis{B}{z}{G}$ and $\axis{B}{x}{H}$ are collinear. This
is only the case if the error quaternion $\tilde{q} \equiv \hat{q}^*q_y$ is in
gimbal lock in terms of ZYX Euler angles. It is important to note that failure
of this method depends only on the \emph{error quaternion}, and not in any way
on the absolute rotations $\hat{q}$ and $q_y$. As a result, the algorithm is
equally stable in all global orientations of the body, as desired.

If the algorithm fails, a backup algorithm that zeros the Euler ZXY yaw of
\fr{H} with respect to \fr{G} is employed instead. Analogously to
\eqnrefs{zyxyawmethodbegin}{zyxyawmethod},
\begin{align}
\axis{B}{y}{H} &= (\hat{x}\hat{y}+\hat{w}\hat{z},\: 0.5-\hat{x}^2-\hat{z}^2,\: \hat{y}\hat{z}-\hat{w}\hat{x}) \\
\axis{B}{\tilde{y}}{G} &= \axis{B}{y}{H} - (\axis{B}{y}{H} \cdot \axis{B}{z}{G})\axis{B}{z}{G} \\
\axis{B}{\tilde{x}}{G} &= \axis{B}{\tilde{y}}{G} \times \axis{B}{z}{G} 
\end{align}
after which equations \eqnrefs{normalisevecs}{vecstorotmat} are used the same as
before. Given that the previous algorithm failed, this algorithm is guaranteed
never to, hence completing the ZYX yaw method.

\subsection{Fused Yaw Orientation Resolution Method}

The fused yaw resolution method is quite similar in idea to the ZYX yaw method,
only instead of zeroing the ZYX yaw of \fr{H} with respect to \fr{G}, it zeros
the fused yaw. The first notable distinction here to before is that having zero
relative fused yaw is in fact a mutual relationship, as the inverse of a
rotation has the exact negative of its fused yaw. The second notable distinction
is that the notion of fused yaw is more closely related to quaternions than ZYX
yaw, and so a convenient direct quaternion formulation exists. Treating
quaternions notationally in \eqnref{quatfmethod} as column vectors in $\R^4$,
the algorithm can be summarised mathematically as
\begin{align}
\axis{H}{z}{G} &= L_{\hat{q}}(\axis{B}{z}{G}) = \hat{q}\smidge\axis{B}{z}{G}\smidge\hat{q}^*
= (z_{Gx},z_{Gy},z_{Gz}) \label{eqn:calcHzG}\\
\rot{\bar{q}}{G}{B}_y &=
\begin{bmatrix}
1+z_{Gz} \nsmidge & -z_{Gy} & z_{Gx} & 0 \\
z_{Gy} & \nsmidge 1+z_{Gz} \nsmidge & 0 & -z_{Gx} \\
-z_{Gx} & 0 & \nsmidge 1+z_{Gz} \nsmidge & -z_{Gy} \\
0 & z_{Gx} & z_{Gy} & \nsmidge 1+z_{Gz}
\end{bmatrix}
\rot{\hat{q}}{G}{E}
\label{eqn:quatfmethod}
\end{align}
where \eqnref{calcHzG} is calculated numerically as in \eqnref{quatrotvec}, and
$\rot{q}{G}{B}_y$ is subsequently calculated as the normalisation of
$\rot{\bar{q}}{G}{B}_y$. The mathematical proof of the correctness of this
algorithm is beyond the scope of this paper. From inspection it can be seen
however, that the only case in which the algorithm fails is
$\axis{H}{z}{G} = (0,0,-1)$. This is the case if the \emph{error quaternion}
$\tilde{q}$ is a rotation by exactly $\pi$ radians about an axis in the
$\vect{x}\vect{y}$ plane. This is a subset of the conditions on $\tilde{q}$
required for unstable equilibrium of the passive filter itself. As such, the use
of the fused yaw resolution method ensures that there is only a single error
condition for which any part of the total passive filter yields suboptimal
results. Furthermore, this one error condition is when $\tilde{q}$ is at an
exact antipode of the identity rotation---a case that in practical situations is
never reached. Nevertheless, for reasons of completeness and robustness, the
above algorithm falls back to zeroing the ZYX yaw if it fails. This is computed
using \eqnrefs{normalisevecs}{zyxyawmethod}, and is guaranteed not to fail if
the fused yaw algorithm failed. It is important to note that this resolution
algorithm is once again equally stable in all global orientations of the body.

\section{Extensions to the Estimator}
\label{sec:extensions}

\subsection{Quick Learning}
\label{sec:quicklearning}

It is desired for the attitude estimator to settle quickly from large estimation
errors, yet simultaneously provide adequate general noise rejection. To this
end, we propose \emph{quick learning} as a method to help achieve this. Quick
learning allows two sets of PI gains to be tuned---one set that provides
suitably fast transient response, and one set that provides good tracking and
noise rejection. Given a desired quick learning time, a parameter
$\lambda \in [0,1]$ is used to fade linearly between the two sets of gains,
ending in the nominal setpoint tracking gains. The gain fading scheme is given by
\begin{equation*}
(k_p,k_i) = \lambda(k_p^{nom},k_i^{nom}) + (1-\lambda)(k_p^{quick},k_i^{quick}).
\end{equation*}
Quick learning can be triggered at any time, including automatically when the
estimator starts, and is disabled when $\lambda$ reaches 1.

\subsection{Estimation with Two-Axis Acceleration Data}
\label{sec:reducedacc}

If only two-axis $\vect{x}\vect{y}$ accelerometer data is available, then the
missing z-component of $\axis{B}{\tilde{a}}{}$ can be calculated by solving
$\norm{\axis{B}{\tilde{a}}{}} = g$, where $g$ is the magnitude of gravitational
acceleration. Letting $\axis{B}{\tilde{a}}{} = (a_x,a_y,a_z)$, this yields
\begin{equation}
a_z = -\sqrt{\max\{g^2-a_x^2-a_y^2,0\}}\,.
\end{equation}
This however, only allows for attitude estimates in the positive z-hemisphere, as
the sign of the missing $a_z$ component has to be assumed. For many applications, like
bipedal walking, this can be sufficient.

\subsection{Estimation with Reduced Magnetometer Data}

Two-axis $\vect{x}\vect{y}$ magnetometer data can still be used for
$\axis{B}{\tilde{m}}{}$ if the third unknown component is left to zero. Due to
the projection operation in \eqnref{magprojection}, this in general still
produces satisfactory results. The calibration process of $\axis{G}{m}{e}$, the
Earth's magnetic field, also remains the same, as the z-component thereof is not
required for the orientation resolution algorithm. If magnetometer data is only
available in terms of relative heading angle $\psi$, then the required
three-axis data can be constructed using $\axis{B}{m}{} = (\cos{\psi},\sin{\psi},0)$,
and used as before.

\subsection{Estimation without Magnetometer Data}
\label{sec:nomag}

If no magnetometer data is available in a system, then the attitude estimator
can still be used without any degradation in the estimation quality in the pitch
and roll dimensions by setting $\axis{B}{m}{}$ and $\axis{G}{m}{e}$ to zero. In
this case, the estimation relies solely on the selected yaw-based orientation
resolution method. Due to the yaw-zeroing approach used, the open-loop yaw
produced by the estimator remains stable with each update of $q_y$. The linear
combination of gyroscope biases that corresponds to rotations in the
instantaneous $\axis{}{x}{G}\axis{}{y}{G}$ plane however does not have feedback,
so small constant global yaw velocities in $\hat{q}$ can result. This yaw drift
is unavoidable as no yaw feedback is present without magnetometer data. A more
stable output quaternion $\hat{q}_s$ can be obtained by removing the fused yaw
component of the estimate. Letting $\hat{q} = (\hat{w},\hat{x},\hat{y},\hat{z})$,
this can be done using
\begin{equation}
\tilde{\hat{q}}_s = (\hat{w},0,0,-\hat{z}) \mspace{3mu} \hat{q}, \quad \hat{q}_s = \frac{\tilde{\hat{q}}_s}{\norm{\tilde{\hat{q}}_s}}.
\end{equation}
We do not recommend removing the ZYX yaw instead, as this leads to
unexpected behaviour near the not uncommon scenario of pitch rotations by
$\tfrac{\pi}{2}$ radians.

\section{Experimental Results}
\label{sec:results}

\begin{figure}[!tb]
\centering
\parbox{\linewidth}{\centering\includegraphics[width=1.0\linewidth]{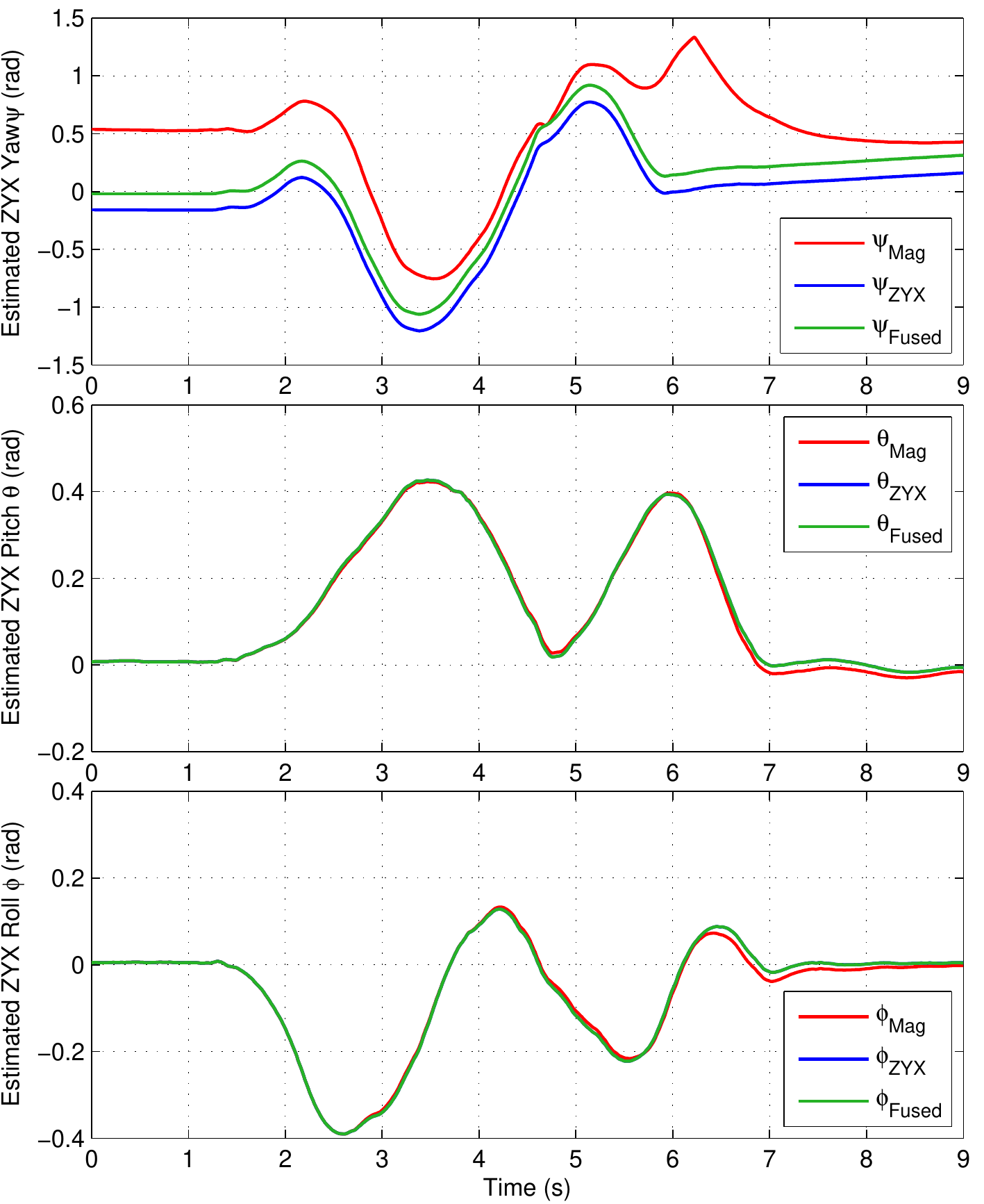}}
\caption{Estimation results on the \nop robot, using the magnetometer, ZYX yaw
orientation resolution, and fused yaw orientation resolution.\vspace{-2em}} %
\label{fig:att_est_results}
\end{figure}

Thorough experimentation and testing of the proposed attitude estimator and
corresponding \cpp implementation has been performed in simulation and on
multiple humanoid robots. Presented in \figref{att_est_results} are the
results of three parallel instances of the attitude estimator running on a \nop
robot. The same L3G4200D gyroscope measurements, LIS331DLH accelerometer
measurements, and HMC5883L magnetometer measurements were made available to each
of the estimators. One of the estimators was permitted to use the magnetometer
data, while the other two were configured to only use the ZYX yaw and fused yaw
orientation resolution methods respectively. It can be seen that the
results of the two yaw resolution methods are virtually indistinguishable, apart
from in the ZYX yaw plot, where there is a vertical shift between the two
curves. This can be expected to happen due to the lack of feedback in the yaw
axis in these two methods. A (more approximate) vertical shift can also be seen
between the non-magnetometer yaw curves and the yaw produced by the magnetometer
method. An important observation is that while the use of the magnetometer
allows the yaw estimation to be meaningful and absolute instead of just
relative, it only minimally affects the pitch and roll, as desired. The effect
on the pitch and roll is not exactly zero though, due to the interplay between
the added yaw information and the measured angular velocities. Note that the
magnetometer chip temporarily provided incorrect values shortly after
\SI{5.0}{\second}. The estimator recovery, and performance in the pitch and roll
directions despite this significant input disturbance, attests to the stability
of the estimator. In applications where a yaw orientation estimate is required,
the magnetometer method is the clear choice, but suitable magnetometer
measurements must be available. Otherwise, the authors recommend the fused yaw
resolution method for mathematical and performance reasons.

\begin{figure}[!tb]
\centering
\parbox{\linewidth}{\centering\includegraphics[width=1.0\linewidth]{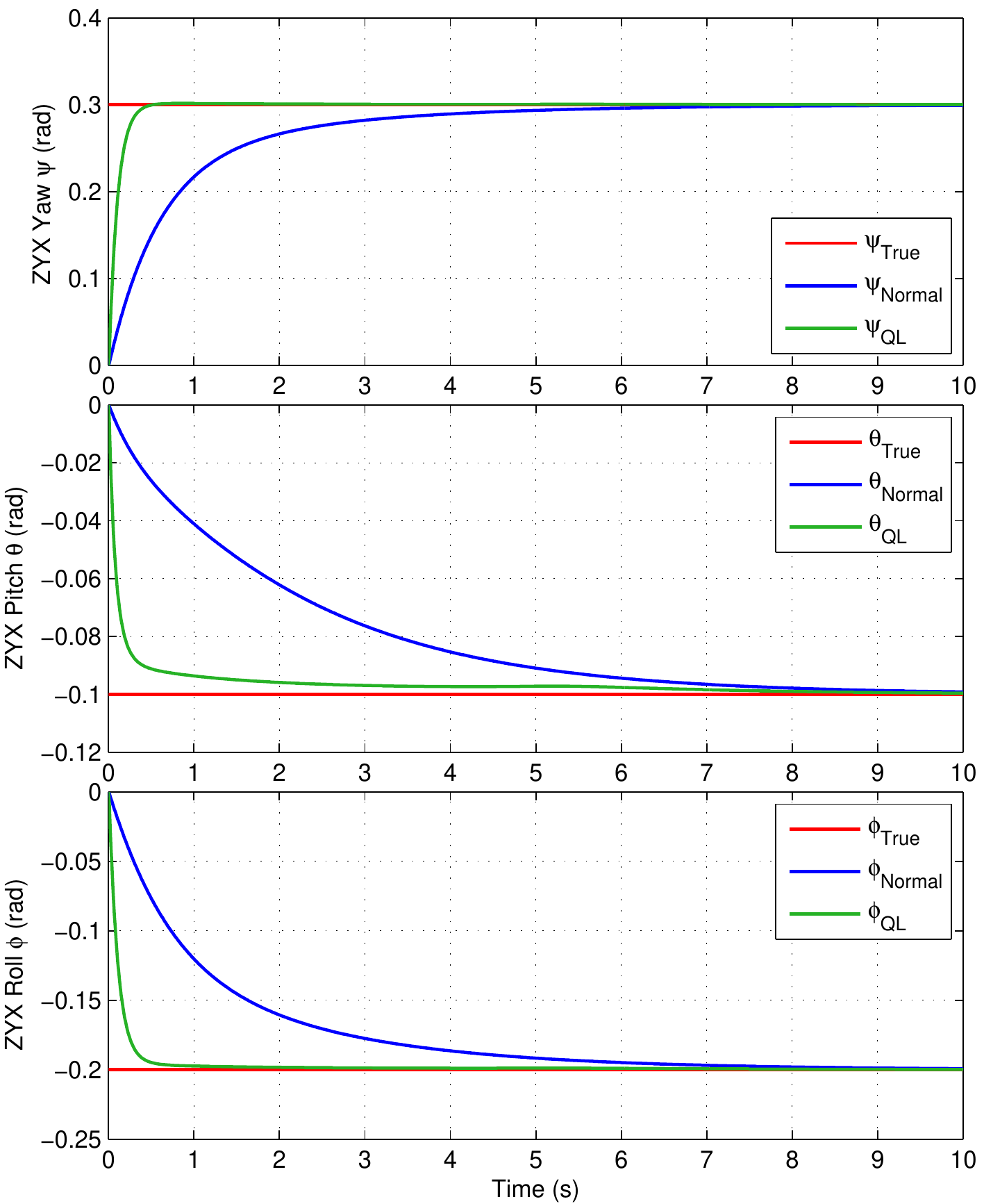}}
\caption{Simulated attitude estimation results demonstrating the effect of quick learning
on the filter's transient response.\vspace{-2em}} %
\label{fig:att_est_ql_results}
\end{figure}

The effect of quick learning on the filter's transient response is shown in
\figref{att_est_ql_results}. For a simulated step in the true orientation of a
body, the response of the filter with and without quick learning activated is
shown. The quick learning time used was \SI{3.0}{\second}. The $\psi_{QL}$,
$\theta_{QL}$ and $\phi_{QL}$ waveforms, using the quick learning feature, show
clear improvement in rise time and settling time over the $\psi_{Normal}$,
$\theta_{Normal}$ and $\phi_{Normal}$ waveforms, for which quick learning was
disabled.

The attitude estimator was designed to be able to run at high loop rates on
embedded hardware, so as to minimise estimation and possible control feedback
latencies. The \cpp attitude estimator library code was tested on a PC with a
\SI{2.40}{\giga\hertz} Intel i5-2430M processor. On a single CPU core, the
average execution time of the estimator over 100 million cycles was found to be
\SI{127.6}{\nano\second} for the magnetometer method, \SI{144.3}{\nano\second}
for the ZYX yaw method, and \SI{112.3}{\nano\second} for the fused yaw method.
It is to be expected that the fused yaw method takes comparatively less time, as
it does not in general require a rotation matrix to quaternion conversion,
unlike the other two methods. From these results it is confidently anticipated
that the algorithm is efficient enough to be implemented at high execution rates
on a low cost microcontroller, where floating and/or fixed point operations are
comparatively more expensive than on a PC.

\section{Conclusions}
A filter for attitude estimation, released online in the form of a \cpp library
\cite{AttEstGithub}, has been presented in this paper. The filter uses the
technique of complementary filtering, and builds on the nonlinear passive
complementary filter presented by Mahoney et al. in \cite{Mahony2008}, providing
robust algorithms for the reconstruction of a `measured' orientation from
instantaneous sensor data. The filter is equally stable in all global body
orientations, and only demonstrates potential non-convergent behaviour on a
pathological set that is of no practical concern. Extensions to the filter allow
for reliable attitude estimation in situations of reduced sensory data, and the
advent of quick learning allows for quicker settling times from large estimation
errors when required. The output of the presented attitude estimator can be
used, for example, for the analysis and control of balance in a biped robot. For
this task it can be beneficial to further decompose the estimated orientation
into its components in each of the major planes. Future work includes a method
for accounting for the inertial components of the accelerometer measurements
based on other sensors and/or system information.

\appendix

This appendix introduces the notation, definitions and well-known identities
that are used throughout this paper. The set of all unit vectors in $\R^3$, the
2-sphere, is denoted $S^2$. The set of all rotation matrices is called the
special orthogonal group $\text{SO}(3)$, and is defined as
\begin{equation*}
\text{SO}(3) = \{R \in \R^{3\times3} \st R^TR=\I, \,\text{det}(R)=1\}.
\end{equation*}
Rotation of a vector $\vect{v} \in \R^3$ by a rotation matrix is given by matrix
multiplication. For a rotation from coordinate frame \fr{A} to frame \fr{B}, we
have that
\begin{equation}
\rot{R}{A}{B} =
\begin{bmatrix}
\axis{A}{x}{B} & \mspace{-5mu}\axis{A}{y}{B} & \mspace{-5mu}\axis{A}{z}{B}
\end{bmatrix}
 = 
\begin{bmatrix}
\axis{B}{x}{A} & \mspace{-5mu}\axis{B}{y}{A} & \mspace{-5mu}\axis{B}{z}{A}
\end{bmatrix}^T,
\label{eqn:rotmatrowcol}
\end{equation}
where for example $\axis{A}{y}{B}$ is the column vector corresponding to the
y-axis of frame \fr{B}, expressed in the coordinates of \fr{A}. The term
$\rot{R}{A}{B}$ refers to the rotation from \fr{A} to \fr{B}.

The set of all quaternions $\Hsym$, and the subset $\Q$ thereof of all
quaternions that represent pure rotations, are defined as
\begin{align*}
\Hsym &= \{q = (q_0,\vect{q}) = (w,x,y,z) \in \R^4\} \\
\Q &= \{q \in \Hsym \st \abs{q} = 1\}.
\end{align*}
The rotation of a vector $\vect{v} \in \R^3$ by a quaternion $q \in \Q$ is given
by the function $L_q(\vect{v}) : \R^3 \rightarrow \R^3$, defined as
\begin{align}
L_q(\vect{v}) &= q\vect{v}\conj{q} \notag\\
&= (q_0^2-\norm{\vect{q}}^2)\vect{v} + 2(\vect{q}\cdot\vect{v})\vect{q} + 2q_0(\vect{q}\times\vect{v}) \notag\\
&= \vect{v} + q_0\vect{t} + \vect{q}\times\vect{t}, \label{eqn:quatrotvec}
\end{align}
where $t = 2(\vect{q}\times\vect{v})$, and $q^*$ is the quaternion conjugate of
$q$. Note that \eqnref{quatrotvec} provides the computationally most efficient
method for calculating $L_q(\vect{v})$. Relative quaternion rotations are
denoted using symbols such as $\rot{q}{A}{B}$.

Two vectors that are linearly dependent and a positive multiple of each other
are referred to as \emph{parallel}. Two linearly dependent vectors that are a
negative multiple of each other are referred to as \emph{antiparallel}. Two
vectors that are either parallel or antiparallel are referred to as \emph{collinear}.

The conversion between the quaternion and rotation matrix representations of a
rotation is often required, but not entirely numerically trivial. Given a unit
quaternion $q = (w,x,y,z) \in \Q$, the equivalent rotation matrix is given by
\begin{equation*}
R =
\begin{bmatrix}
1-2(y^2+z^2) & 2(xy-wz) & 2(xz+wy) \\
2(xy+wz) & \mspace{-6.0mu}1-2(x^2+z^2)\mspace{-6.0mu} & 2(yz-wx) \\
2(xz-wy) & 2(yz+wx) & 1-2(x^2+y^2)
\end{bmatrix}
\end{equation*}
Depending on how the rotation matrix $R$ is subsequently used, it may be
necessary to coerce each of the matrix entries to $[-1,1]$. Although for
$\abs{q} = 1$ it is impossible in a mathematical sense for one of the entries to
exceed unity in magnitude, it can happen in practice due to floating point
arithmetic. In such cases, subsequent calculations such as $\alpha=\acos(R_{33})$
can lead to unwanted numerical problems.

The reverse conversion is more difficult and is split into four cases, where
each case corresponds to one of the four quaternion parameters being taken as
the base of the conversion. Given a rotation matrix $R \in \text{SO}(3)$ with
matrix entries $R_{ij}$, if $\tr(R) \geq 0$,
\begin{align*}
r &= \sqrt{1 + R_{11} + R_{22} + R_{33}} \\
q &= \Bigl( \tfrac{1}{2}r, \tfrac{1}{2r}(R_{32}-R_{23}), \tfrac{1}{2r}(R_{13}-R_{31}), \tfrac{1}{2r}(R_{21}-R_{12}) \Bigr)
\end{align*}
Else if $R_{33} \geq R_{22}$ and $R_{33} \geq R_{11}$,
\begin{align*}
r &= \sqrt{1 - R_{11} - R_{22} + R_{33}} \\
q &= \Bigl( \tfrac{1}{2r}(R_{21}-R_{12}), \tfrac{1}{2r}(R_{13}+R_{31}), \tfrac{1}{2r}(R_{32}+R_{23}), \tfrac{1}{2}r \Bigr)
\end{align*}
Else if $R_{22} \geq R_{11}$,
\begin{align*}
r &= \sqrt{1 - R_{11} + R_{22} - R_{33}} \\
q &= \Bigl( \tfrac{1}{2r}(R_{13}-R_{31}),\tfrac{1}{2r}(R_{21}+R_{12}),\tfrac{1}{2}r,\tfrac{1}{2r}(R_{32}+R_{23}) \Bigr)
\end{align*}
And otherwise,
\begin{align*}
r &= \sqrt{1 + R_{11} - R_{22} - R_{33}} \\
q &= \Bigl( \tfrac{1}{2r}(R_{32}-R_{23}),\tfrac{1}{2}r,\tfrac{1}{2r}(R_{21}+R_{12}),\tfrac{1}{2r}(R_{13}+R_{31}) \Bigr)
\end{align*}
This implementation of the rotation matrix to quaternion conversion is extremely
robust, as it always chooses as the base of the conversion the quaternion
parameter that provides the most well-conditioned problem to solve.

\bibliographystyle{IEEEtran}
\bibliography{IEEEabrv,ms}
\end{document}